\documentclass[conference]{IEEEtran}
\IEEEoverridecommandlockouts
\usepackage{cite}
\usepackage{amsmath,amssymb,amsfonts}
\usepackage{algorithmic}
\usepackage{graphicx}
\usepackage{textcomp}
\usepackage{xcolor}
\usepackage{acronym}
\usepackage{subcaption}
\def\BibTeX{{\rm B\kern-.05em{\sc i\kern-.025em b}\kern-.08em
    T\kern-.1667em\lower.7ex\hbox{E}\kern-.125emX}}

\begin{document}



\title{Towards spiking analog hardware implementation of a trajectory interpolation mechanism for smooth closed-loop control of a spiking robot arm}

\DeclareRobustCommand{\IEEEauthorrefmark}[1]{\smash{\textsuperscript{\footnotesize #1}}}

\author{\IEEEauthorblockN{Daniel Casanueva-Morato\IEEEauthorrefmark{1,2},
Chenxi Wu\IEEEauthorrefmark{3,4}, Giacomo Indiveri\IEEEauthorrefmark{4},\\
Juan P. Dominguez-Morales\IEEEauthorrefmark{1,2} and
Alejandro Linares-Barranco\IEEEauthorrefmark{1,2}}
\IEEEauthorblockA{\IEEEauthorrefmark{1}Robotics and Technology of Computers Lab., ETSII-EPS, Universidad de Sevilla, Sevilla, Spain}
\IEEEauthorblockA{\IEEEauthorrefmark{2}Smart Computer Systems Research and Engineering Lab (SCORE), I3US, Universidad de Sevilla, Spain}
\IEEEauthorblockA{\IEEEauthorrefmark{3}SynSense AG Corporation, Zurich, Switzerland}
\IEEEauthorblockA{\IEEEauthorrefmark{4}Institute of Neuroinformatics, University of Zurich and ETH Zurich, Switzerland, Spain\\dcasanueva@us.es}}

\maketitle

\begin{abstract}
  Neuromorphic engineering aims to incorporate the computational principles found in animal brains, into modern technological systems.
  Following this approach, in this work we propose a closed-loop neuromorphic control system for an event-based robotic arm.
  The proposed system consists of a shifted Winner-Take-All spiking network for interpolating a reference trajectory and a spiking comparator network responsible for controlling the flow continuity of the trajectory, which is fed back to the actual position of the robot.
  The comparator model is based on a differential position comparison neural network, which governs the execution of the next trajectory points to close the control loop between both components of the system.
  To evaluate the system, we implemented and deployed the model on a mixed-signal analog-digital neuromorphic platform, the DYNAP-SE2, to facilitate integration and communication with the ED-Scorbot robotic arm platform.
  Experimental results on one joint of the robot validate the use of this architecture and pave the way for future neuro-inspired control of the entire robot.
\end{abstract}

\begin{IEEEkeywords}
event-based robot, trajectory interpolation, spiking neural networks, DYNAP-SE2, neuromorphic control
\end{IEEEkeywords}

\acrodef{SNN}[SNN]{Spiking Neural Network}
\acrodef{ED}[ED]{Event-Driven}
\acrodef{ED-ScorBot}[ED-Scorbot]{Event-Driven ScorBot}
\acrodef{DoF}[DoF]{degrees of freedom}
\acrodef{AER}[AER]{Address-Event-Representation}
\acrodef{SPID}[SPID]{Spike-based PID}
\acrodef{HDL}[HDL]{Hardware Description Language}

\section{Introduction}

The brain is capable of efficiently solving complex problems in real-time and with low energy consumption \cite{sun2023brain, vaz2023backbone}. This is why neuromorphic engineering focuses on the development of systems that attempt to mimic the brain \cite{mead1990neuromorphic, indiveri2011neuromorphic} in order to incorporate its superior computational capabilities into current systems. For this purpose, neuromorphic systems use \acp{SNN} for their implementations.

\acp{SNN} are a type of neural network that mimics the way biological brains process and transmit information by means of electrical impulses, called spikes. Unlike traditional neural networks, \acp{SNN} use learning mechanisms based on local synaptic plasticity \cite{khacef2023spike}. Systems implemented with such networks therefore are characterized by asynchronous and distributed operation \cite{kim2020spiking}.

In robotics, one of the most common tasks involves executing trajectories to reach a specific target in space \cite{dai2022review}. Currently, the movements of these systems are primarily governed by adaptive differential control \cite{xu2023design, trivedi2023biomimetic}, which emphasizes precise and efficient execution, particularly in space but also in time \cite{liu2023robotic}. Nonetheless, such movements, and consequently the trajectories, frequently exhibit oscillatory (sometimes erratic) behavior around the target point or non-uniform behavior in space within a continuous trajectory \cite{rosmann2012trajectory, rosmann2015planning}. At the same time, human movements are characterized by being non-efficient but uniform and distributed, reaching the target position in a safer, smoother and more natural manner \cite{zhao2023human, prada2013dynamic, linares2022towards}.
By emulating the neural dynamics of biological systems, neuromorphic systems presents significant potential in their application to robotic trajectory control.

Other studies were presented in the literature that implement the control of robotic arms using neuromorphic controllers to benefit from the advantages inherent to these systems \cite{Stagsted_Sandamirskaya_Loihi_PID_IROS_2020, Stroonbants_PID_MAVs_Loihi_ICONS2022, zhao_risi_jetcas_2020_pid_icub}. Notable among these are \cite{casanueva2024integrating, linares2022towards}, where a neuromorphic hardware platform was used to control a robotic arm in the spiking domain. However, these works implement an open-loop trajectory control system that requires some form of intervention prior to or during the execution of trajectories, and additionally, the trajectories over time are executed in a stepped manner in space, meaning they are non-uniform.

Considering these limitations, this work proposes a closed-loop neuromorphic control system for a robotic arm. The system performs trajectory interpolation based on the reference trajectory to achieve continuous and smooth motion over time. This system connects the Event-Driven Scorbot (ED-Scorbot) robotic platform \cite{edscorbot} to a \ac{SNN} deployed on the hybrid analog-digital hardware platform DYNAP-SE2 \cite{richter2024dynap}. The proposed neural network is based on a shifted WTA network for the interpolation of the trajectory provided to the robotic platform, combined with a differential position comparison neural network, which enables commanding next trajectory points to close the control loop between both components of the system. The \ac{AER} protocol is used to encode and communicate neuronal activity from the neuromorphic processor chip to the \ac{SPID} controllers operating in parallel for the 6 \ac{DoF} of this robot. Experiments on a joint of the robot validate the use of this architecture and pave the way for future neuro-inspired control of the entire robot.

\section{Materials and methods}

\subsection{\textit{DYNAP-SE2 neuromorphic platform}}

The proposed neural network was developed using \acp{SNN} that are compatible with the constraints and features of neuromorphic hardware. To evaluate the network and to test it in a real robotic setup, the model was implemented and deployed on the hybrid analog-digital neuromorphic hardware platform DYNAP-SE2\cite{richter2024dynap}. The platform employs a scalable multicore design that features various memory architectures to enable dynamic asynchronous processing of events and integrates analog circuits that replicate synaptic and neural behaviors with digital circuits responsible for managing network connections and spike routing between active neurons. The neuron circuits implement a model equivalent to the AdExp-I\&F~\cite{Brette_Gerstner05}, and the synapse dynamics are implemented using the current-mode Differential Pair Integrator log-domain filter~\cite{bartolozzi2007synaptic}, allowing 4 possible synapse types: AMPA, NMDA, GABA and SHUNT. The circuits in each core, which encode parameters such as neuron leakage and refractory period, are designed with the same nominal values. However, due to device mismatch, the actual values for each circuit may vary. DYNAP-SE2 circuits typically exhibit a coefficient of variation of approximately 20\% \cite{zendrikov2023brain}. Consequently, there can be significant discrepancies between the specified and actual values of these parameters.

DYNAP-SE2 communication with external platforms is done through asynchronous messages using the \ac{AER} protocol. Each position of the robot is identified by the activation of a specific neuron, i.e., a one-hot encoding. This encoding offers benefits such as increased robustness in the representation of information, greater ease in identifying and comparing positions, and simplification of the proposed neural network.

\subsection{\textit{Spiking Neural Network}}

The network receives as input the reference position of the trajectory to be executed and the robot's real-time position. It outputs the adjusted reference position to smooth the robot's movement, along with a signal indicating that the actual reference position has been reached, allowing the next reference position in the trajectory to be provided. To achieve this smoothing, the network performs interpolation of the reference position that needs to be reached, adding an offset to the actual position. The robotic arm will not stop or oscillate around the actual target position due to its adaptive control; instead, by setting the reference further ahead, it will go through without stopping. Just as it passes the actual reference position of the trajectory, the network will provide the next reference position with an offset again, causing the robot to stop only at a change of direction or at the end of the trajectory. 

To achieve this functionality, the network is divided into a shifted WTA and a comparator. Furthermore, the robot's position is decomposed into two parts: the most significant information about the position will be the coarse grain, while the least significant information will be the fine grain. The former will be used for interpolating and smoothing the trajectory, as well as determining when the robot has reached the actual reference position, while the latter will be employed to more accurately and proactively prevent when the robot is about to reach the actual reference position.

\subsubsection{\textit{Shifted WTA}} 
\begin{figure}[ht]
 \centering
    \includegraphics[width=1.0\linewidth]{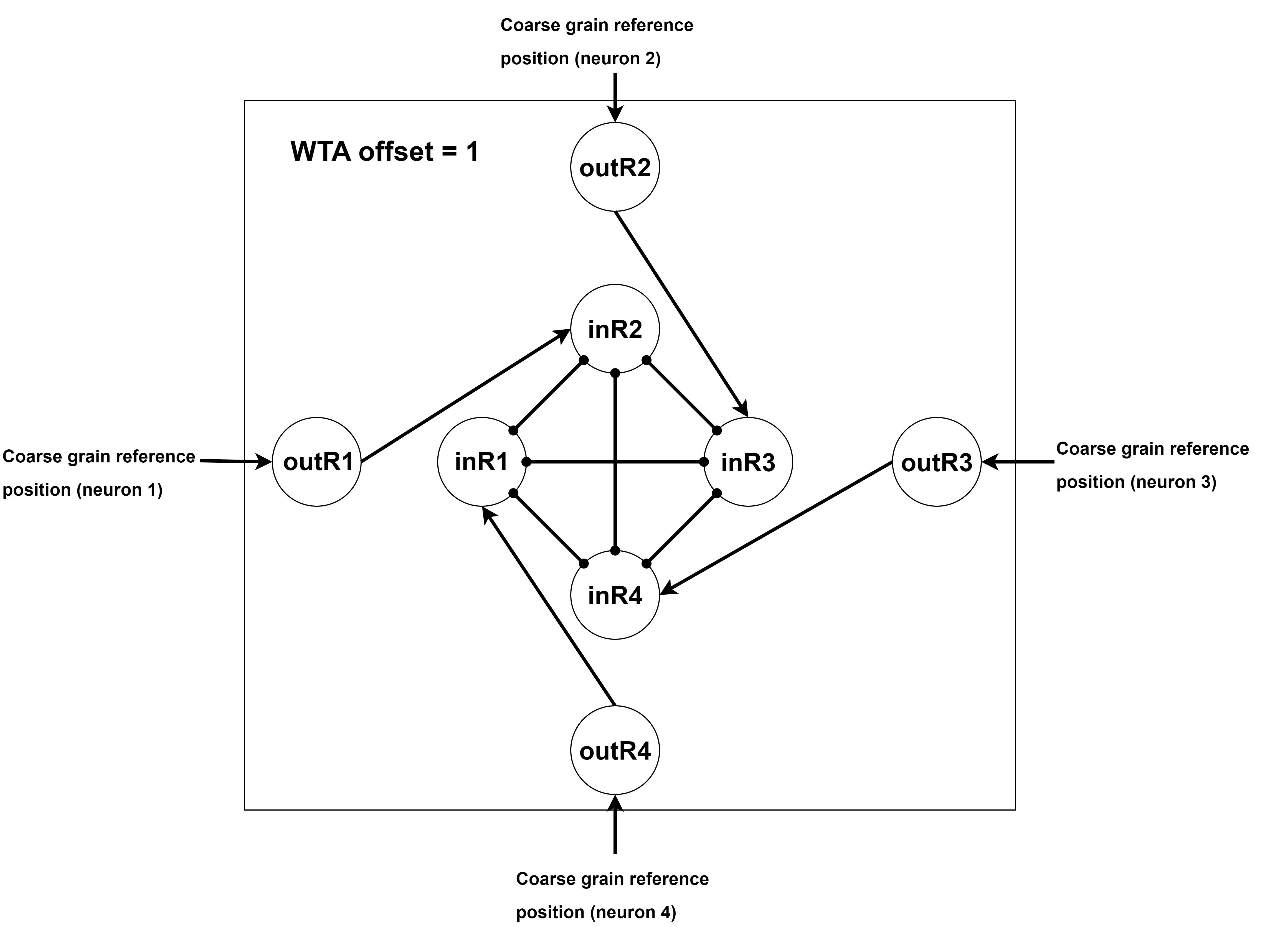}
 \caption{Shifted WTA SNN architecture. Lines with arrow or circular ends represent excitation with AMPA or inhibition with SHUNT dendritic channels, respectively. }
  \label{fig:wtaarch}
\end{figure}

It is responsible of interpolating the reference positions. It consists of a WTA network with an external (outer) ring of neurons as the input layer ($outR$) connected to an internal (inner) ring of neurons as the output layer ($inR$). For each possible position to be interpolated (\textit{pRef\_i}), there is a corresponding neuron in the external ring ($outR\_i$) that identifies it and will be activated at the input of the network. The difference with conventional WTA architectures is that each neuron in the external ring does not connect to its equivalent ($i$) in the internal ring ($inR\_i$), but rather connects to the $i + x \% numPosRef$, where $i$ is the index of the reference position, $x$ is the offset of the interpolation, and $numPosRef$ is the number of reference positions considered by the network. Given an input reference position, it returns that position with a configurable offset $x$. The architecture of this SNN is illustrated in Fig.\ref{fig:wtaarch}, where an example with 4 possible positions and an offset of 1 is considered.

\subsubsection{\textit{Comparator}} 
\begin{figure}[!ht]
    \centering
        \begin{subfigure}[b]{\linewidth}
            \includegraphics[width=\linewidth]{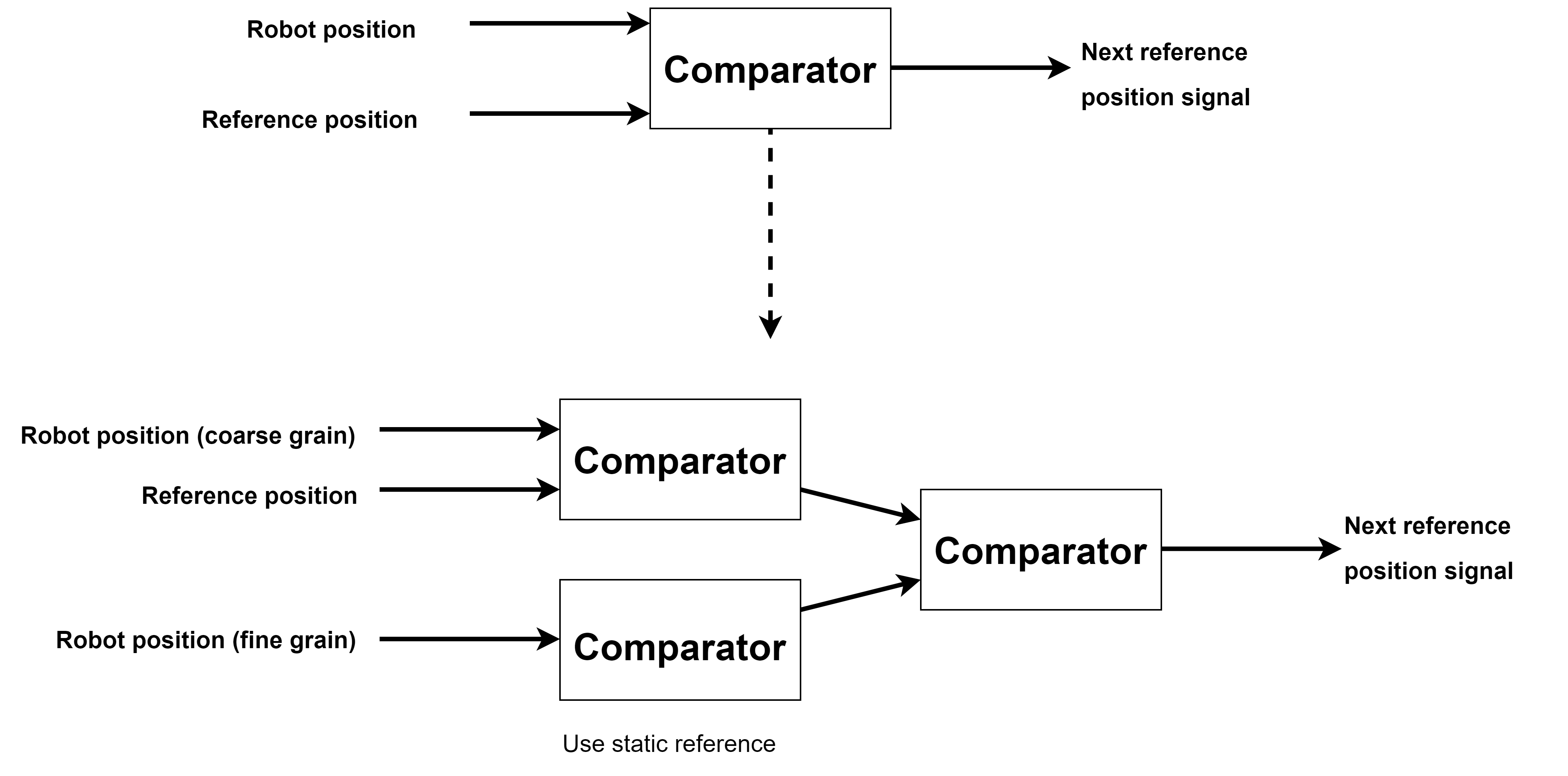}
            \caption{}
        \end{subfigure}
        \begin{subfigure}[b]{0.55\linewidth}
            \includegraphics[width=0.95\linewidth]{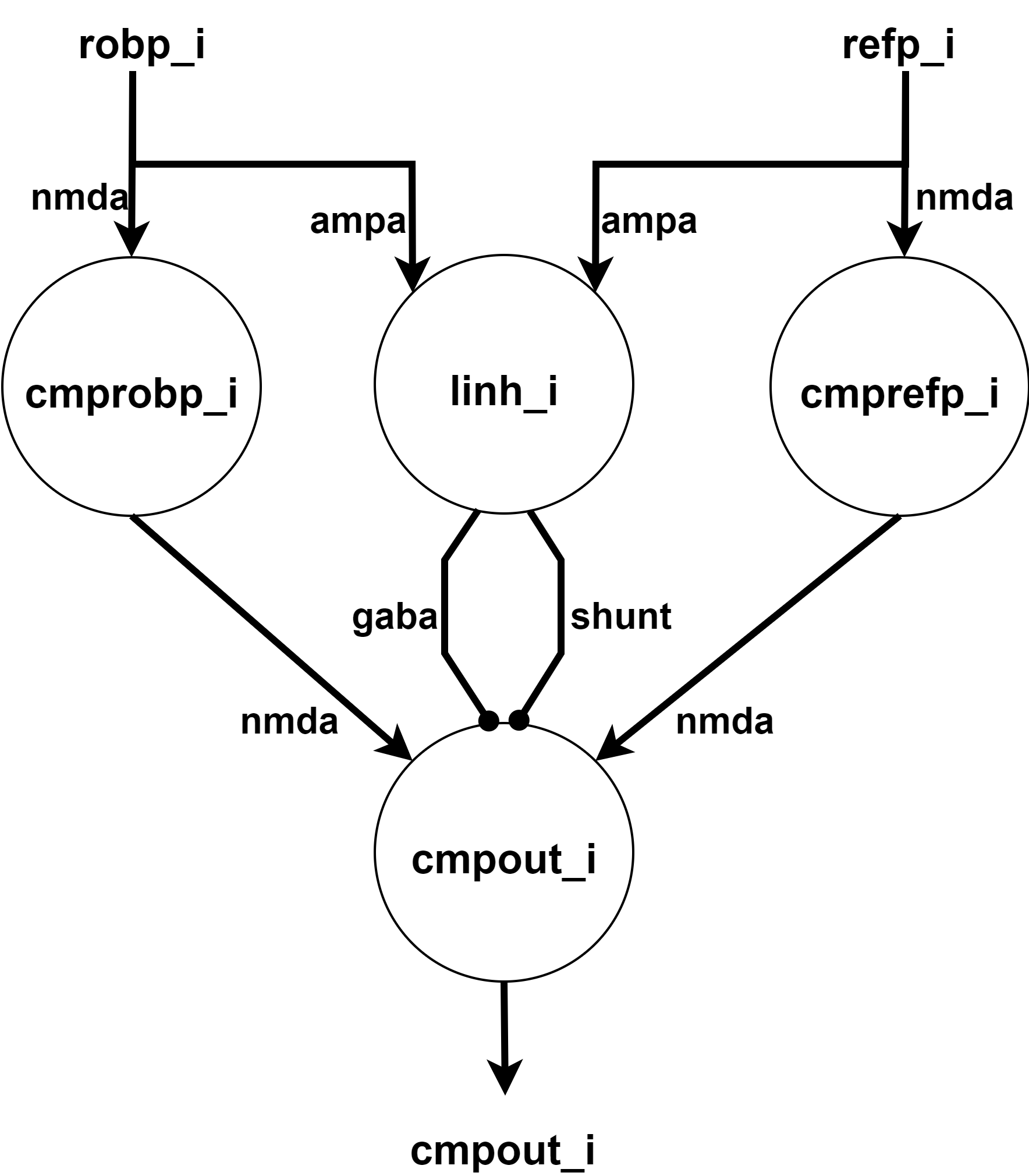}
            \caption{}
        \end{subfigure}
        \caption{SNN comparator architecture (a) at neural block level and (b) at computational neural units level.}
        \label{fig:cmparch}
\end{figure}

It is the SNN responsible for determining when the robot's joint has reached the real reference position. It receives the robot's position and the actual reference position, generating an output signal indicating when both positions match, and consequently, the next reference position in the trajectory can be received for interpolation. This network consists of 3 subnetworks (Fig.\ref{fig:cmparch}a). The first network is responsible for coarse grain position comparison to determine when the destination is reached. The second network is dedicated to comparing the robot's fine-grain position with a static reference, configurable by design, providing greater precision and anticipation to the system to determine the next reference position. The third network takes both output signals from the previous two networks, activating its output when both coincide. This last one is responsible for generating the output signal of the system to request the next reference position.


Thanks to the encoding of position information in one-hot, the design of each of these networks is made up of small computational (neural) comparison units (Fig.\ref{fig:cmparch}b). Each of these units allows individual comparison of a specific input position A with another specific input position B. In this way, possible input positions are compared in pairs independently and in parallel. The operation of this unit is similar to a logical AND gate, when both inputs activate within a time window, the output is activated. When working with spike trains at the input, the output (\textit{cmpout\_i}) will only be activated when both excitatory input neurons (\textit{cmprobp\_i} and \textit{cmprefp\_i}) overcome the inhibitory input train (\textit{linh\_i}). The GABA channel is used to maintain constant inhibition over the input activity time, while the SHUNT channel is responsible for small inhibition primarily focused at the beginning of input activity to prevent possible unwanted activations of the output as a result of a burst of potential coming from only one of the inputs.


\subsection{\textit{Event-Driven Scorbot}}

The robotic arm used for this work, \ac{ED-ScorBot}, is a ScorBot-ER VII model \cite{Scorbot_user_manual} with six \ac{DoF} that has been modified to work with an \ac{ED} motor controller \cite{edscorbot}. A Zynq 7100 PSoC from Xilinx hosts the VHDL description of an adapted \ac{SPID} motor controller \cite{spid_sensors} for joint position control with pulse-frequency-modulation (PFM)\cite{linares2006aer, linares2020ed}. Those FPGA-based controllers can convert the received position reference into a spike train \cite{aer_gen_exha_iwann05} and then use spike-based processing building blocks \cite{ssp_building_blocks} to design the controller. The \ac{ED-ScorBot} logic has been improved to allow the monitoring of internal spiking information with the use of an \ac{AER} monitor \cite{usbaermini2} and to allow the use of \ac{AER}-based spiking inputs for generating the position references of the joints from external \ac{AER} systems, like Loihi \cite{davies2018loihi}, SpiNNaker \cite{furber2014spinnaker}, Reckon \cite{linaresbarranco2024adaptiveroboticarmcontrol} or DYNAP-SE2 \cite{richter2024dynap} \cite{linares2022towards}.

\subsection{\textit{DYNAP-SE2 and ED-Scorbot interface}}

\begin{figure}[ht]
 \centering
    \includegraphics[width=1.0\linewidth]{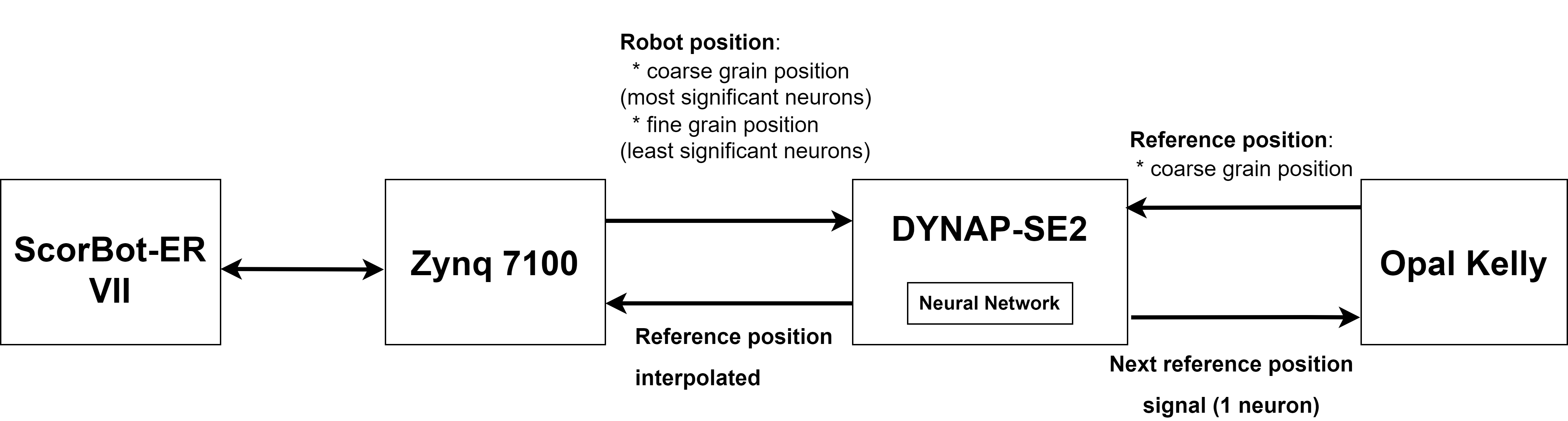}
 \caption{Diagram of the information flow and connection of the robotic arm and the DYNAP-SE2 platform.} 
  \label{fig:conn}
\end{figure}

\begin{figure}[ht]
 \centering
    \includegraphics[width=0.9\linewidth]{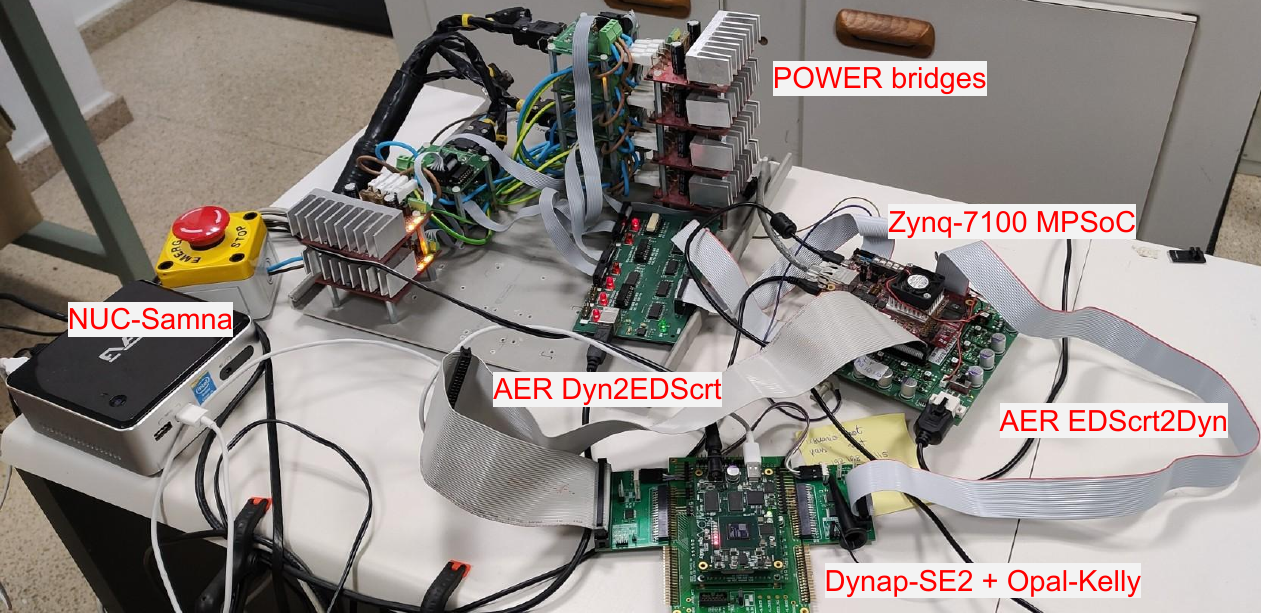}
 \caption{Robot controller and DYNAP-SE2 in loop-back.}
  \label{fig:hardwaresetup}
\end{figure}

Fig.\ref{fig:conn} shows the connectivity needed for the communication between both hardware platforms. DYNAP-SE2 and the robotic arm are indirectly connected. On the one hand, the ED-Scorbot is connected to an FPGA board (Zynq 7100), which is responsible for the event-based control and for managing its input/output information. This FPGA board will handle the communication with DYNAP-SE2 directly. On the other hand, DYNAP-SE2 is connected to a second FPGA board. This second board is responsible for deploying the proposed neural networks on DYNAP-SE2, as well as providing an interface for a simulated spiking input and output information. In this work, an Opal Kelly board is used, which will be responsible for sending the event-encoded information regarding the reference trajectory to DYNAP-SE2. Fig.\ref{fig:hardwaresetup} depicts the hardware setup necessary for communication.

\subsection{Experiments}

\begin{figure}[!ht]
    \centering
    \includegraphics[width=0.7\linewidth]{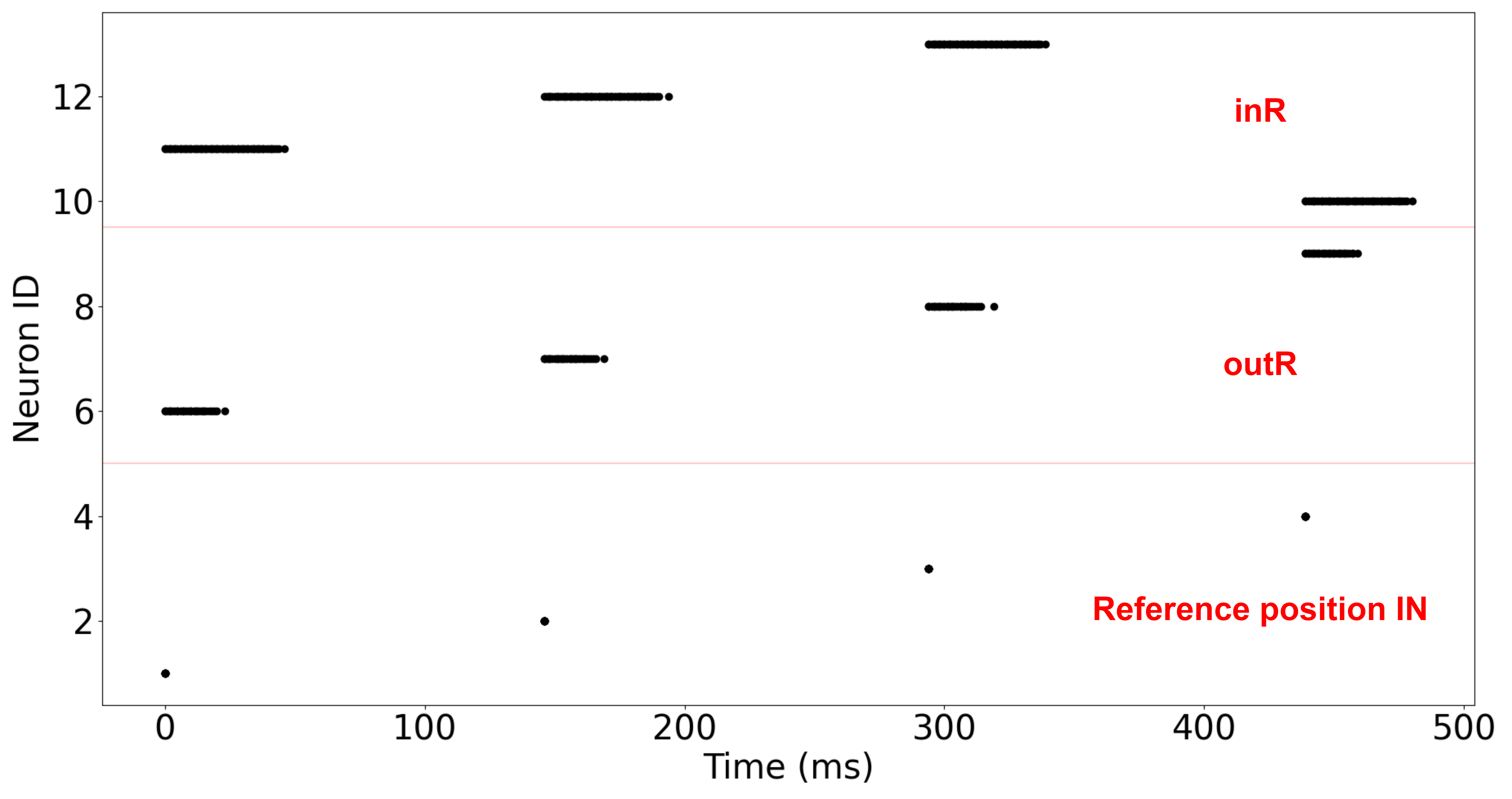}
    \caption{Raster plot of the spiking activity of the shifted SNN WTA during the sweep of all possible inputs for 4 possible reference positions for 1 joint with an offset of 1.}
    \label{fig:wta}
\end{figure}

\begin{figure}[ht]
 \centering
    \includegraphics[width=0.8\linewidth]{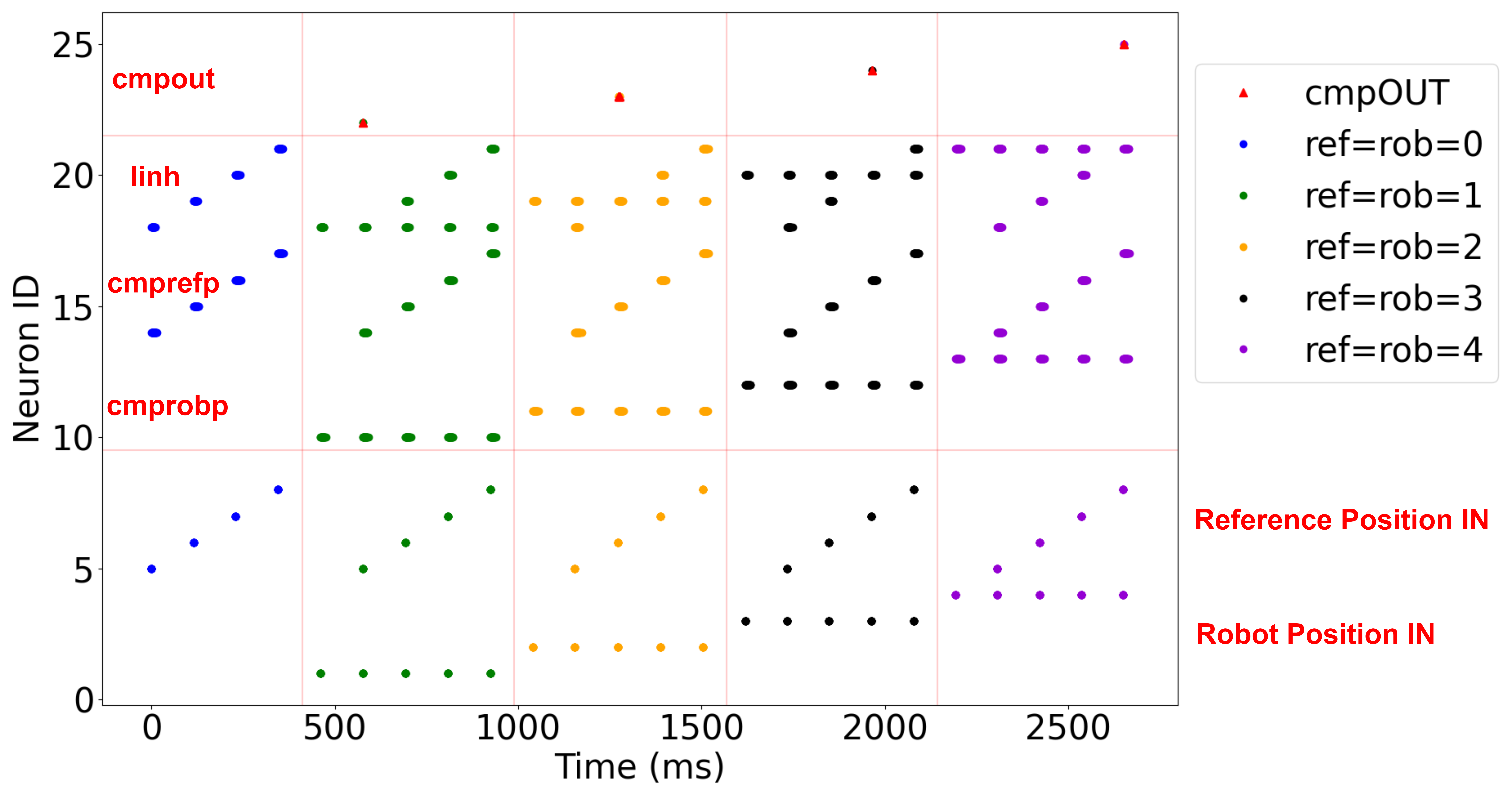}
 \caption{Raster plot of the spiking activity of the SNN comparator during the sweep of all possible input positions for 4 input robot and reference positions.}
  \label{fig:cmp}
\end{figure}

The proposed system has been evaluated through a set of experiments. Using the internal event generators of the DYNAP-SE2 platform, the input and output interactions of the SNN with the robotic platform were simulated. Fig.\ref{fig:wta} shows the spiking activity of the shifted WTA network when stimulated by a sweep of all possible input combinations for the case of 4 reference positions in a joint with an offset of 1. Fig.\ref{fig:cmp} shows the spiking activity of the coarse grain comparator network when stimulated by a sweep of all possible combinations of 4 reference positions and 4 coarse grain robot positions. The activity displayed in both figures verifies the correct functioning of both networks separately; in the shifted WTA network, it is observed that the input activity to the outer ring of the network is shifted to its output with the correct offset in the inner ring, while the comparator network shows that only in cases where the activation of the same robot position and reference coincides, the output signal is activated.

\begin{figure}[ht]
 \centering
    \includegraphics[width=0.99\linewidth]{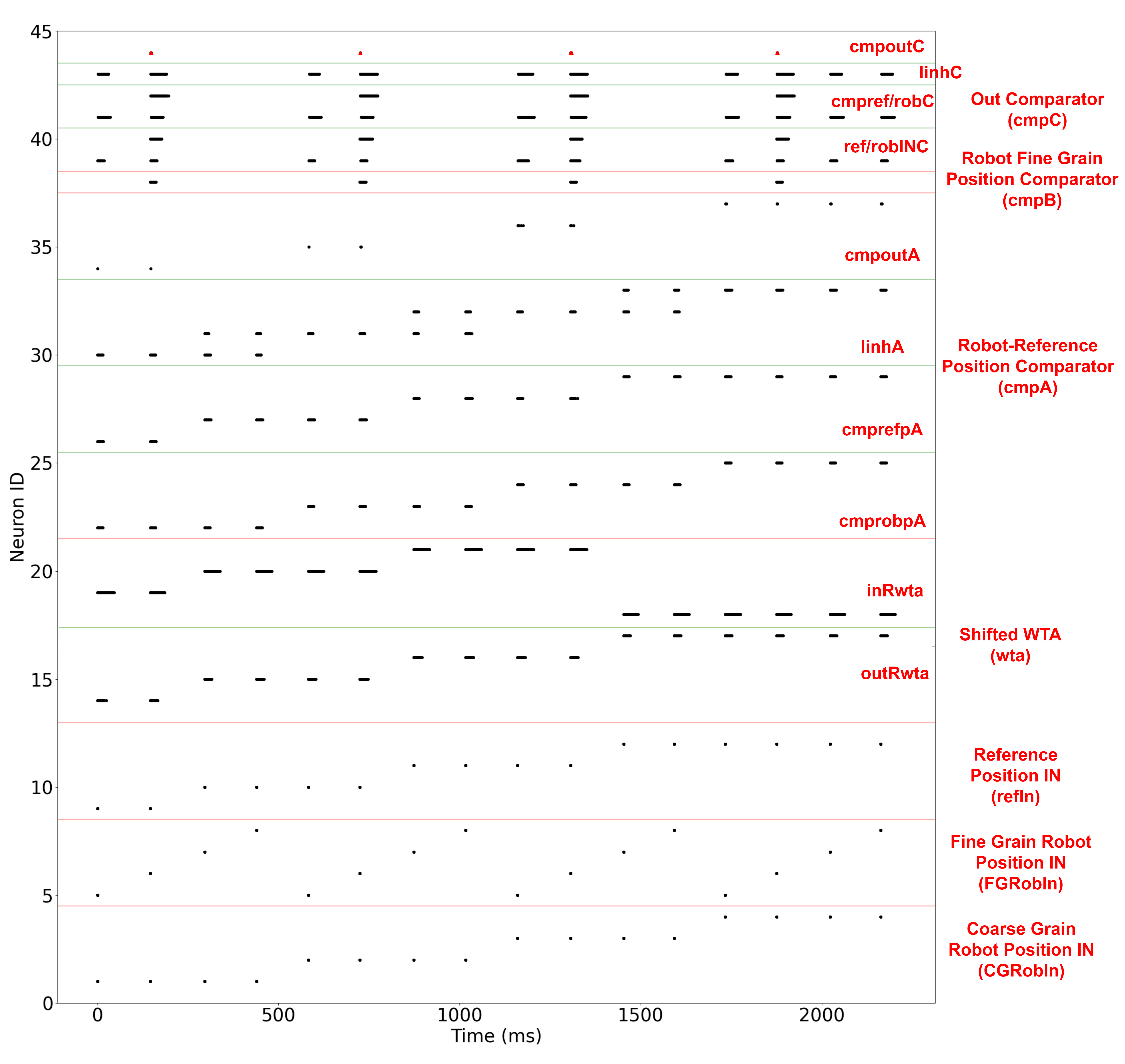}
 \caption{Raster plot of the spiking activity of the full SNN network during the operation of interpolation of the ascending trajectory (from the closest to the furthest position from the home position) of one of the robot joints with a resolution of 4 positions in coarse and fine grain.}
  \label{fig:fullnetwork}
\end{figure}

Fig. \ref{fig:fullnetwork} shows the result of stimulating the network with an ascending trajectory input for 1 joint with 4 possible positions. The trajectory starts at position 0 and progresses until it reaches position 3. Throughout the path, it can be seen how the shifted WTA network correctly interpolates the reference position and how the comparator is capable of accurately measuring when the actual reference position was reached, activating the output signal and prompting the input of a new reference position interpolated by the shifted WTA network.

\section{Conclusions}


In this work, we proposed an event-based neuromorphic closed-loop robotic control system, for trajectory interpolation and smoothing. The system consists of a shifted WTA spiking network for interpolating the reference trajectory and a comparator spiking network responsible for controlling the flow continuity of the trajectory, which is fed back with the actual position of the robot. Both networks have been implemented and deployed on the DYNAP-SE2 analog-digital neuromorphic processor to facilitate integration and communication with the ED-Scorbot robotic platform. This hardware platform manages the different joints of a robotic arm using a SPID and is powered by the output of the SNN.

Through a series of experiments, the proper functioning of the various components defining the SNN has been verified, not only in isolation but also in conjunction. Furthermore, the complete hardware setup has been deployed, establishing the communication and connection protocol between DYNAP-SE2 and the robotic platform, ensuring a closed-loop structure. Future work will focus on the experimentation of the complete hardware setup. Due to the limitations of the DYNAP-SE2 chip, this has only been demonstrated for a set of 4 possible positions in both coarse and fine granularity. Future work includes modifications and extensions of the network to enable independent control of each of the 4 available joints in the robotic arm. It would also be interesting to connect DYNAP-SE2 to a neuromorphic memory model deployed on another platform responsible for supplying the various reference positions of the trajectory to be interpolated, achieving a fully neuromorphic system. This memory model could be implemented on a hybrid platform like DYNAP-SE \cite{Casanueva_Morato_2024} or even on a digital platform such as SpiNNaker \cite{casanueva2024bio}.

\section*{Acknowledgment}



This research was partially supported by NEKOR (PID2023-149071NB-C54) from MICIU/AEI /10.13039/501100011033 and by USECHIP (TSI-069100-2023-001) of PERTE Chip Chair program, funded by European Union – Next Generation EU. D. C.-M. was supported by a FPU Scholarship from the Spanish MICIU.



\bibliographystyle{IEEEtran}
\bibliography{bibliography}

\end{document}